\documentclass[twocolumn, switch]{article} 

\usepackage{preprint}

\usepackage[numbers,square]{natbib}
\usepackage[utf8]{inputenc}	
\usepackage[T1]{fontenc}	
\usepackage{xcolor}		
\usepackage[colorlinks = true,
            linkcolor = purple,
            urlcolor  = blue,
            citecolor = cyan,
            anchorcolor = black]{hyperref}	
\usepackage{booktabs} 		
\usepackage{nicefrac}		
\usepackage{microtype}		
\usepackage{lineno}		
\usepackage{float}			

\usepackage{lipsum}		

\usepackage{newfloat}
\DeclareFloatingEnvironment[name={Supplementary Figure}]{suppfigure}
\usepackage{sidecap}
\sidecaptionvpos{figure}{c}

\usepackage{amsmath}
\usepackage{amssymb}
\usepackage{mathtools}
\usepackage{amsthm}

\usepackage{listings}
\usepackage{xcolor}

\lstdefinelanguage{yaml}{
  keywords={true,false,null,y,n},
  keywordstyle=\color{blue}\bfseries,
  basicstyle=\ttfamily\small,
  comment=[l]{\#},
  commentstyle=\color{gray},
  stringstyle=\color{teal},
  moredelim=[l][\color{orange}]{:},
  sensitive=false
}
\usepackage{titlesec}
\titlespacing\section{0pt}{12pt plus 3pt minus 3pt}{1pt plus 1pt minus 1pt}
\titlespacing\subsection{0pt}{10pt plus 3pt minus 3pt}{1pt plus 1pt minus 1pt}
\titlespacing\subsubsection{0pt}{8pt plus 3pt minus 3pt}{1pt plus 1pt minus 1pt}

\usepackage{tikz,xcolor,hyperref}

\definecolor{lime}{HTML}{A6CE39}
\DeclareRobustCommand{\orcidicon}{
	\begin{tikzpicture}
	\draw[lime, fill=lime] (0,0) 
	circle [radius=0.16] 
	node[white] {{\fontfamily{qag}\selectfont \tiny ID}};
	\draw[white, fill=white] (-0.0625,0.095) 
	circle [radius=0.007];
	\end{tikzpicture}
	\hspace{-2mm}
}
\foreach \x in {A, ..., Z}{\expandafter\xdef\csname orcid\x\endcsname{\noexpand\href{https://orcid.org/\csname orcidauthor\x\endcsname}
			{\noexpand\orcidicon}}
}

\title{$n$-Musketeers: Reinforcement Learning Shapes Collaboration Among Language Models}
\newcommand\titleheader{A preprint template for LaTeX users (Header)}

\usepackage{xwatermark}

\usepackage{authblk}

\author[1\thanks{\texttt{rmasukaw@uci.edu}}]{Ryozo Masukawa}
\author[1]{Sanggeon Yun}
\author[1]{Hyunwoo Oh}
\author[1]{SungHeon Jeong}
\author[1]{Raheeb Hassan}
\author[1]{Hanning Chen}
\author[1]{Wenjun Huang}
\author[4]{Mahdi Imani}
\author[2]{Pietro Mercati}
\author[3]{Nathaniel D. Bastian}
\author[1]{Mohsen Imani}

\affil[1]{Department of Computer Science, University of California, Irvine}
\affil[2]{Intel Corporation}
\affil[3]{United States Military Academy at West Point}
\affil[4]{Department of Electrical and Computer Engineering, Northeastern University}

\begin{document}

\twocolumn[ 
  \begin{@twocolumnfalse} 
  
\maketitle

\begin{abstract}

Recent progress in reinforcement learning with verifiable rewards (RLVR) shows that small, specialized language models (SLMs) can exhibit structured reasoning without relying on large monolithic LLMs.
We introduce \textbf{soft hidden-state collaboration},
where multiple heterogeneous frozen SLM experts are integrated through their internal representations via a trainable attention interface.
Experiments on Reasoning Gym and GSM8K show that this latent integration is competitive with strong single-model RLVR baselines.
Ablations further reveal a dual mechanism of expert utilization: for simpler arithmetic domains, performance gains can largely be explained by static expert preferences, whereas more challenging settings induce increasingly concentrated and structured expert attention over training, indicating emergent specialization in how the router connects to relevant experts.
Overall, hidden-state collaboration provides a compact mechanism for leveraging frozen experts, while offering an
observational window into expert utilization patterns and their evolution under RLVR.

\end{abstract}
\vspace{0.35cm}

  \end{@twocolumnfalse} 
] 



\vspace{-10mm}
\section{Introduction}
\vspace{-3mm}
\label{sec:intro}


Recent large language models (LLMs), exemplified by DeepSeek-R1~\cite{guo2025deepseek}, demonstrate that reinforcement learning with verifiable rewards (RLVR), together with distillation, can transfer strong multi-step reasoning behaviors into more compact language models (LMs)\footnote{Unless specified otherwise, we use the term ``LM'' to refer to autoregressive decoder-only language models in this paper.}.
In particular, these results suggest that small language models (SLMs) ($\sim$10B parameters) can retain such reasoning competence under substantial compression, with some reaching or surpassing GPT-3.5-class performance on reasoning benchmarks~\cite{slm3_abdin2024phi}.
These findings indicate that structured reasoning ability can be induced through outcome-level supervision, rather than being inherently tied to model scale.
If reasoning competence can indeed be transferred across models of different sizes, the design space for practical agentic systems shifts from scaling monolithic LLMs toward composing collections of off-the-shelf expert SLMs~\cite{slm_future}. A natural next question is how such heterogeneous experts can be combined within a single trainable policy, without retraining the experts themselves or relying on discrete text-based aggregation.
This shift has motivated a growing body of work on multi-LM reasoning systems that coordinate multiple frozen models~\cite{wu2024autogen,hong2024metagpt,multi_agent_ex1}.
\begin{figure}[t!]
    \centering
    \includegraphics[width=\linewidth]{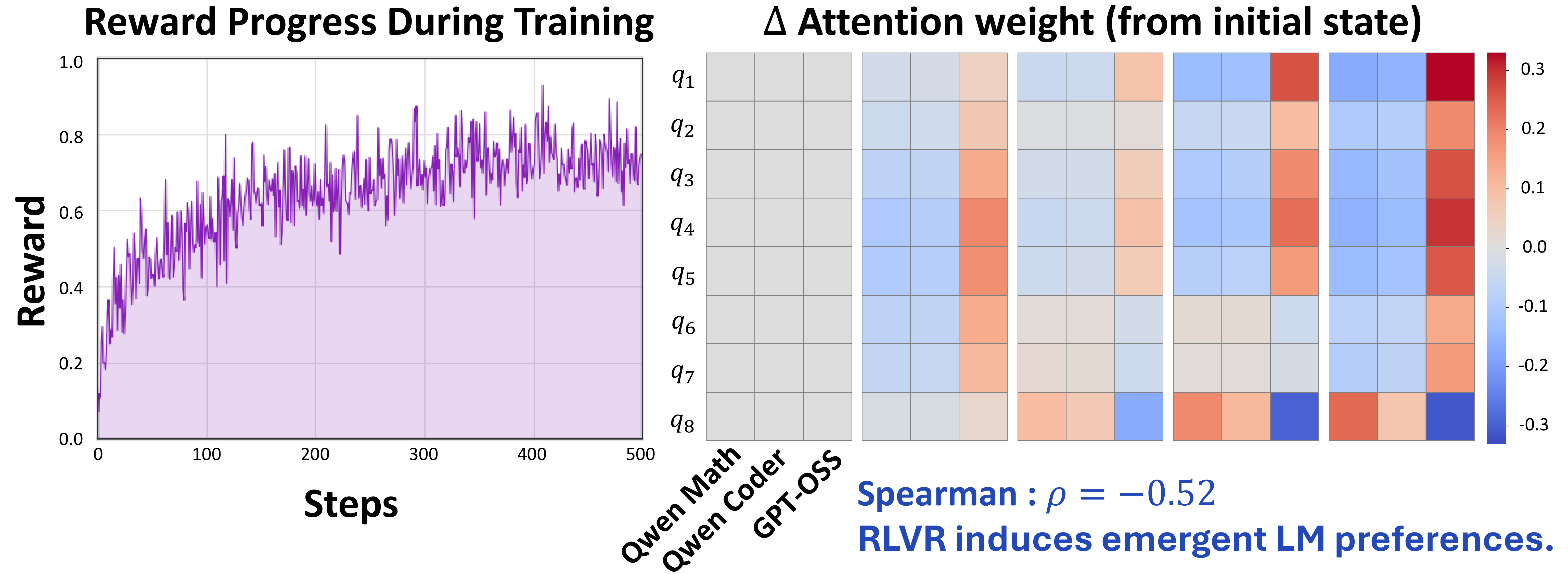}
    \caption{During training with RLVR, expert LM utilization becomes increasingly structured, without routing or expert-role supervision.
Left: reward during training. Right: change in expert attention from initialization, illustrating emergent expert roles.
}
    \vspace{-10mm}

    \label{fig:honto_teaser}
\end{figure}

\begin{figure*}[t!]
    \centering
    \includegraphics[width=\linewidth]{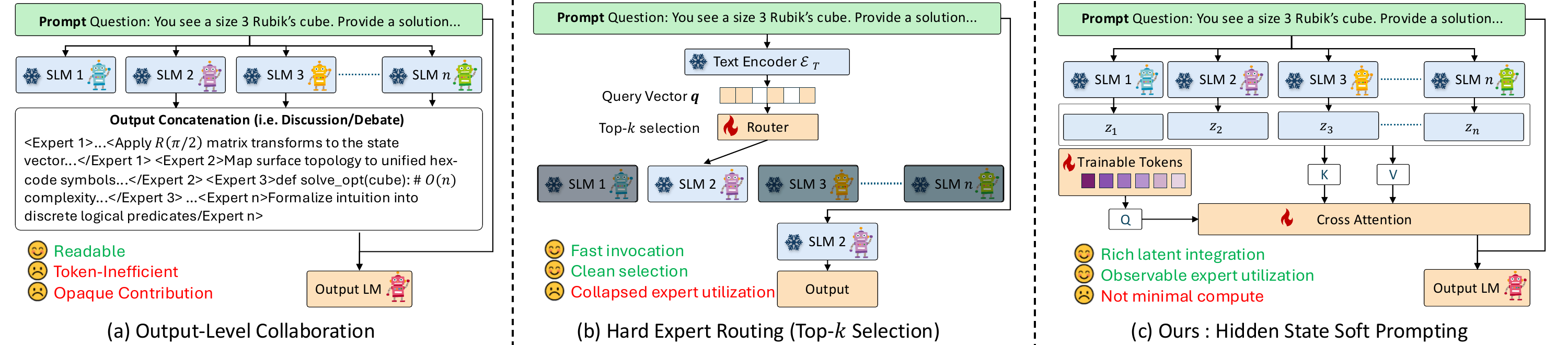}
    \caption{Comparison of (a) output-level textual collaboration, (b) hard expert routing, and (c) our hidden state soft prompting approach, highlighting differences in computational efficiency, expert utilization, and observability of expert utilization.}
    \label{fig:teaser}
    \vspace{-5mm}
\end{figure*}

However, a basic question remains poorly understood:
\textit{Do frozen, off-the-shelf LMs' hidden states actually provide useful information to policies trained purely with RLVR?}
Since expert utilization itself is never directly supervised under RLVR, it is unclear what information the policy learns to extract from expert models, or under what conditions such information is beneficial.
Addressing this question requires mechanisms that expose and condition on expert representations during learning, rather than treating experts as black-box generators.
Most existing multi-LM approaches are not designed to expose how expert information is utilized during learning.
Instead, prior work has largely focused on model-level coordination which can be broadly grouped into two paradigms (\autoref{fig:teaser}.a, b).

Output-level collaboration (\autoref{fig:teaser}.a) combines LLMs through discrete textual communication~\cite{debate,llm_blender,huang2024ensemble,co-llm,wang2025mixtureofagents,chen2024frugalgpt}.
While this approach is human-readable and flexible, it is token-heavy and provides only indirect evidence of how individual experts contribute to final decisions.
Hard LM selection via routing (\autoref{fig:teaser}.b) selects one or a small number of experts conditioned on the input~\cite{li2025rt,selection1,selection2,selection3,routier_r1,router_cost_aware_contrastive,router_cook2025brainlike}, improving computational efficiency at the cost of brittle decisions and limited cross-expert interaction.
Architecture-level mixture-of-experts (MoE) methods integrate experts within a unified model~\cite{switchtransformer,heteromoe, router_cook2025brainlike}, but typically require pretraining-scale integration, often training from scratch, and hinder the plug-and-play reuse of heterogeneous, independently developed experts.

To study expert utilization under RLVR, we introduce \textbf{soft hidden state collaboration} (\autoref{fig:teaser}.c),
which integrates multiple frozen expert SLMs via their hidden state representations by generating trainable context tokens in a simple and plug-and-play manner,
allowing arbitrary numbers and combinations of off-the-shelf experts to be incorporated without modifying the experts themselves~\cite{lester-etal-2021-power,li-liang-2021-prefix}. This representation-level adapter exposes expert information as a latent interface to the policy, making expert utilization directly observable during RLVR training.
We implement this adapter with a lightweight Perceiver-style cross-attention module~\cite{jaegle2021perceiver}.
This representation-level adapter allows expert information to influence the policy continuously, while making expert utilization directly observable during training.

We evaluate this approach on Reasoning Gym~\cite{reasoninggym} and GSM8K~\cite{cobbe2021gsm8k}, where soft hidden-state collaboration improves reasoning performance over strong baselines, with gains varying across tasks and expert teams and reaching up to 22.9\% in the best-performing cases.
Beyond accuracy, we observe structured correlations between reward progression and expert utilization, indicating that expert hidden states provide actionable auxiliary signals under RLVR, while controlled ablations suggest that a substantial fraction of these gains can be attributed to prefix-style latent conditioning rather than fine-grained expert reasoning traces.
Crucially, even without explicit supervision over LM usage, RLVR induces structured, task-dependent expert utilization that reveals emergent, reward-aligned roles.
The Perceiver-based integration thus serves mainly as a principled interface for observing these dynamics, rather than as a mechanism required for accuracy.


In summary, our contributions are as follows:

\begin{itemize}
    \item We introduce a simple, plug-and-play hidden state collaboration framework that exposes SLM utilization dynamics under outcome-level supervision without relying on hard routing or textual aggregation.
    \item We empirically demonstrate that hidden states from frozen experts, including compact SLMs, contain actionable reasoning signals that can be exploited by RLVR-trained policies.
    \item Through controlled experiments and ablations, we characterize when and how expert representations improve, degrade, or fail to benefit reasoning performance.
\end{itemize}

\section{Problem Formulation}

\label{sec:problem}
Post-training of pretrained LLMs is commonly performed either via supervised fine-tuning (SFT) or RL with task-level feedback such as RLVR. The two-stage paradigm of first applying SFT to establish a strong initial policy and then continuing with RLVR remains a dominant and widely recognized post-training strategy in the reasoning literature~\cite{sft_to_rlvr}. Recent studies widely suggest that RLVR improves mathematical and analytical reasoning performance in LLMs~\cite{guo2025deepseek,rlvr_support_2,rlvr_support3}, although some works remain cautious about interpreting these gains as the acquisition of fundamentally new reasoning capabilities, instead attributing them to a reshaping of existing behaviors under sparse outcome-based supervision~\cite{yue2025does,rlvr_support_1}. In this work, we focus on the RLVR setting and study how a trainable policy can leverage SLMs through a latent interface.

\vspace{-2mm}
\subsection{Frozen off-the-shelf expert LMs.}
\vspace{-2mm}
\label{subsec:expert}
Let $\mathcal{F} = \{LM_1, \dots, LM_{n}\}$ denote the set of frozen expert LMs.
Given an input prompt $x$, the $i$-th frozen expert $LM_i \in \mathcal{F}$ processes $x$ using its own tokenizer $\tau_i$ and architecture, producing a sequence of final-layer hidden states
\begin{equation}
\label{eq:expert_hidden_pf}
H_i^{(L_i)} = (h_{i,1}^{(L_i)}, \dots, h_{i,T_i}^{(L_i)}) = LM_i(\tau_i(x)),
\end{equation}
where $T_i$ denotes the number of tokens produced by $\tau_i$ and $L_i$ indicates the number of hidden layers in $LM_{i}$.
All expert parameters remain fixed during training.
For convenience, we denote the collection of expert representations $\{ H_i^{(L_i)} \}_{i=1}^{n}$ as $\mathcal{H}$.

To incorporate frozen expert information, we introduce an auxiliary set of expert-conditioned context tokens constructed from the aggregated expert representations $\mathcal{H}$:
\vspace{-2mm}
\begin{equation}
\label{eq:context_tokens}
\mathcal{C} = (c_1, \dots, c_{m}) = g_{\phi}(\mathcal{H}),
\end{equation}
\vspace{-7mm}

where $m$ is the number of context tokens. The mapping $g_{\phi}$ may be instantiated differently across collaboration paradigms. For example, in output-level collaboration (\autoref{fig:teaser}.a), expert representations are first transformed into logits via each expert’s LM head and subsequently aggregated through decoding-based procedures. In this case, $g_{\phi}$ is a fixed, non-trainable mapping that is not optimized under the downstream learning objective. Accordingly, the parameters $\phi$ may be either fixed or trainable, depending on how $g_{\phi}$ is instantiated across different baselines. Our method parameterizes $g_{\phi}$ as a trainable latent adapter that produces learnable context tokens injected as prefix conditioning to the policy, allowing their representations to be shaped indirectly through RLVR training. This distinction enables learning not only the final policy, but also how expert information is internally represented and utilized.
\vspace{-2mm}

Inspired by~\cite{lester-etal-2021-power, li-liang-2021-prefix}, 
we define the expert-conditioned policy as
\begin{equation}
\label{eq:policy_cond_context}
\pi_{\theta,\phi}(y \mid x)
:= \pi_{\theta}\bigl(y \mid [\,x \,\Vert\, \mathcal{C}\,]\bigr),
\end{equation}

where $[\cdot \,\Vert\, \cdot]$ denotes token-sequence concatenation.

Crucially, expert utilization is not directly supervised: there exists no ground-truth annotation specifying what $\mathcal{C}$ should encode, which parts of expert information should be relied upon, or how information from different experts should be combined.
Accordingly, the semantics of $\mathcal{C}$ are defined only through their effect on the policy's behavior.

\vspace{-3mm}

\subsection{RLVR enables end-to-end expert utilization.}
Standard SFT based on Maximum Likelihood Estimation (MLE) optimizes only the conditional likelihood of target outputs.
In our setting, this implies that the context tokens $\mathcal{C}=g_{\phi}(\mathcal{H})$ are never directly supervised and are constrained only indirectly through output likelihood.
As a result, MLE provides no explicit preference for what $\mathcal{C}$ should encode or how expert information should be utilized, beyond what is necessary to match the target outputs.

We therefore adopt RLVR, where the policy is optimized using outcome-level task feedback rather than token-level supervision.
This is particularly suitable in our setting, since the context tokens $\mathcal{C}=g_{\phi}(\mathcal{H})$ are not associated with any ground-truth annotations and their usefulness can only be judged by downstream task success.

Under RLVR, each sampled response $y$ is evaluated by a task-level reward function $r(x,y)$, and the policy is optimized to maximize expected reward over the entire generation process.
Consequently, the construction parameters $\phi$ are updated through their contribution to task outcomes, enabling end-to-end learning of how frozen expert representations should be transformed into context tokens.
A range of RL algorithms, including PPO~\cite{ppo}, GRPO~\cite{guo2025deepseek}, DAPO~\cite{yu2025dapo}, and REINFORCE++~\cite{hu2025reinforce++}, can be viewed as instances of the following KL-regularized expected reward objective:
\begin{equation}
\begin{aligned}
\label{eq:rlvr_obj}
\arg\max_{\theta,\phi}  
\;&
\mathbb{E}_{x\sim\mathcal{D},\, y\sim\pi_{\theta}(\cdot\mid [x \,\Vert\, \mathcal{C}])}
\Big[
r(x,y)
\\
&\quad-
\beta^{-1}
\log
\frac{
\pi_{\theta}\!\big(y\mid [x \,\Vert\, \mathcal{C}]\big)
}{
\pi_{\text{ref}}(y\mid x)
}
\Big].
\end{aligned}
\end{equation}
Here, $\beta>0$ controls the strength of KL regularization to the reference policy $\pi_{\text{ref}}(y\mid x)$, which is the frozen initial policy used to initialize $\pi_{\theta}$.
Thus, RLVR provides a direct learning signal for optimizing expert utilization through the context tokens $\mathcal{C}$.

\vspace{-3mm}
\section{Method: Perceiver-based SLM Adapter}
\vspace{-3mm}

\begin{figure}[t]
    \centering
    \includegraphics[width=\linewidth]{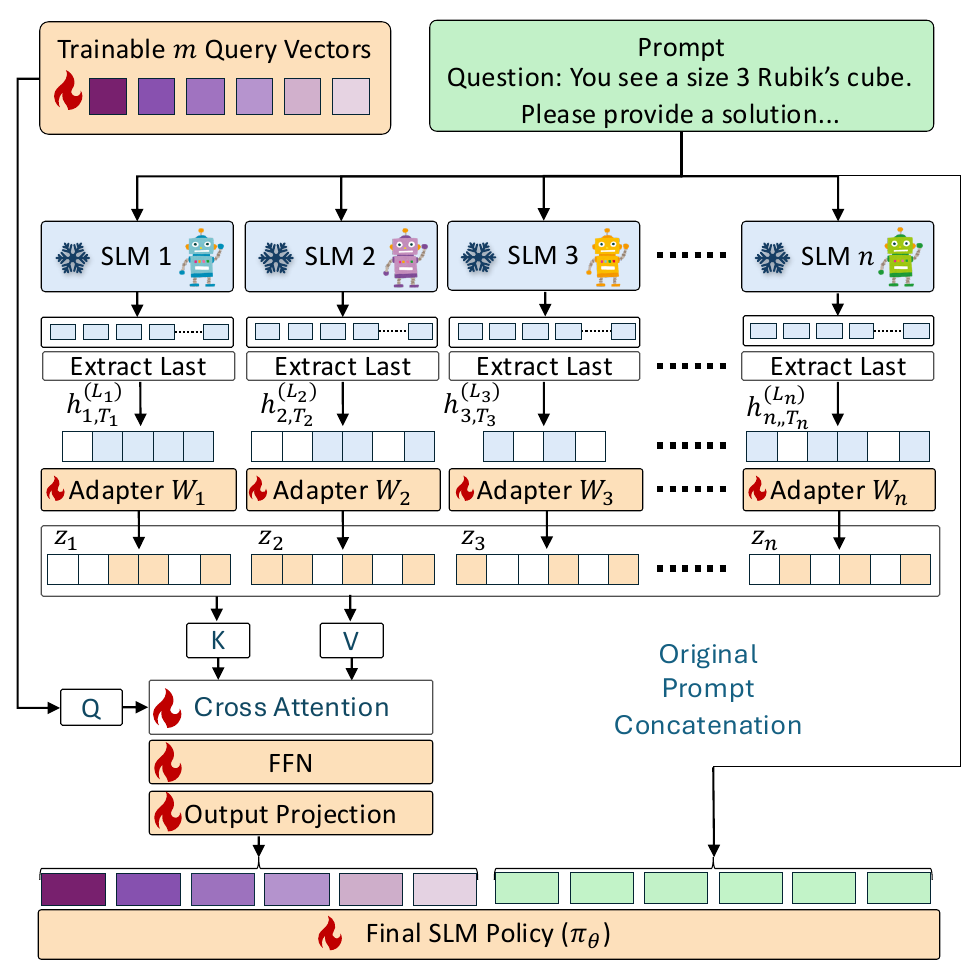}
    \caption{Perceiver-based hidden state collaboration framework.
Frozen heterogeneous SLM experts expose final-layer hidden states, which are aligned and aggregated via trainable latent query tokens using cross-attention. The resulting context tokens condition the final policy model trained with RLVR.}
\vspace{-5mm}
    \label{fig:detailed_method}
\end{figure}

\label{sec:method}
We consider a setting in which multiple heterogeneous, off-the-shelf LMs are available as frozen experts.
Our goal is to integrate their internal representations into a single trainable policy, without retraining the experts or relying on discrete text-based aggregation.
An overview of the proposed framework is shown in ~\autoref{fig:detailed_method}. As introduced in \autoref{sec:problem}, we abstract expert integration as a context-token construction operator $g_{\phi}$ that maps frozen expert representations to context tokens, i.e., $\mathcal{C}=g_{\phi}(\mathcal{H})$. Here, $\phi$ collectively denotes all trainable parameters introduced in this section.

\vspace{-3mm}
\subsection{Expert hidden state extraction and alignment}
\vspace{-2mm}

We follow the frozen expert setup in \autoref{subsec:expert} and use the same notation for expert hidden representations as in \autoref{eq:expert_hidden_pf}.
To obtain a compact prompt-conditioned expert representation, we apply a pooling operator $\mathrm{pool}(\cdot)$ to the expert hidden states and define
\begin{equation}
    \label{eq:last_token}
    h_i = \mathrm{pool}(H_i) = h_{i,T_{i}}^{(L_{i})} \in \mathbb{R}^{d_i},
\end{equation}
where $h_{i,T_{i}}^{(L_{i})}$ denotes the final-layer hidden states of $i$-th expert for an input of length $T_i$, and $d_i$ is the expert-specific hidden dimensionality.
In our default configuration, we use last-token pooling.
This choice is motivated by the autoregressive structure of causal LMs: the final-token hidden state summarizes the full input prefix and parameterizes the conditional next-token distribution. In our framework, experts are evaluated in a forward-only manner; no decoding or sampling is performed.

Under the assumption of heterogeneous experts, we align their representations using expert-specific linear projections.
Let $d$ denote the shared latent dimension.
The aligned expert representations are given by
\begin{equation}
    \label{eq:dim_adapter}
    z_i = h_{i} W_i, \quad W_i \in \mathbb{R}^{d_i \times d}.
\end{equation}
We collect the projected expert representations into an ordered latent sequence
\begin{equation}
    \label{eq:hidden_state}
    V = ( z_1, z_2, \dots, z_n) \in \mathbb{R}^{n \times d}.
\end{equation}

\vspace{-4mm}
\subsection{Latent queries via a Perceiver-style bottleneck}
\vspace{-2mm}

To aggregate the aligned expert representations into a fixed-size auxiliary context, we adopt a Perceiver-style cross-attention mechanism~\cite{jaegle2021perceiver}.
Specifically, we introduce a set of $m$ trainable latent query vectors, stacked as
\begin{equation}
    \label{eq:latent_q}
    Q_{\mathrm{lat}} = ( q_1, \dots, q_m ) \in \mathbb{R}^{m \times d},
\end{equation}
where $m$ is a hyperparameter controlling the size of the latent bottleneck.
Fixing $m$ decouples the number of experts from downstream computation, yielding a constant-size representation regardless of ensemble size.

The interaction between latent queries and expert representations is implemented via scaled dot-product cross-attention.
We first form queries, keys, and values using learned linear projections:
\begin{equation}
\begin{cases}
Q =  Q_{\mathrm{lat}} W_Q , \\
K =  V W_K , \\
V' =  V W_V,
\end{cases}
\end{equation}
where $W_Q, W_K, W_V$ are trainable projection matrices.
We first compute intermediate latent representations via cross-attention followed by a token-wise feed-forward network with $\tanh$ activation (FFN : Appendix \ref{app:ffn}):
\begin{equation}
\label{eq:attention}
\mathcal{C}'
= \mathrm{FFN}\!\left(
Q_{\mathrm{lat}} + \mathrm{softmax}\!\left( \frac{Q K^\top}{\sqrt{d}} \right) V'
\right).
\end{equation}
The final context tokens are then obtained via a linear projection to the policy embedding space:
\begin{equation}
\label{eq:context_tokens_construct}
\mathcal{C}
=
 \mathcal{C}^{\prime}W_{\mathrm{out}} + b_{\mathrm{out}}
\in \mathbb{R}^{m \times d_{\mathrm{out}}},
\end{equation}
where $d_{\mathrm{out}}$ matches the token embedding dimension of the final policy model $\pi_{\theta}$. We initialize the output projection parameters ($W_{out} \in \mathbb{R}^{d\times d_{out}}, b_{out} \in \mathbb{R}^{1\times d_{out}}$) to be near-zero for the sake of stability in early training steps.

Finally, the context tokens $\mathcal{C}$ are provided to the final policy $\pi_\theta$ via prefix-style conditioning, consistent with the expert-conditioned policy definition in \autoref{sec:problem}.
We optimize the trainable parameters in our method using GRPO under the RLVR objective in \autoref{eq:rlvr_obj}.

\vspace{-4mm}
\section{Experiments}
\vspace{-3mm}

\label{sec:exp}
This section empirically evaluates hidden state collaboration under RLVR through a series of controlled comparisons.
Our experiments are designed to answer two questions: (i) whether latent expert tokens improve RLVR performance beyond single-model training, and (ii) whether RLVR induces structured expert roles without routing supervision.
To this end, we systematically vary the design of the context-token construction operator $g_{\phi}$, the expert pool composition, and the source of expert hidden states, while keeping the downstream policy and RLVR setup fixed.




\vspace{-3mm}
\subsection{Experimental Setup}
\vspace{-2mm}

\noindent \textbf{Implementation Details.}
We employ Qwen-2.5-3B-Instruct~\citep{qwen2_5} as the policy $\pi_{\theta}$, with trainable LoRA adapters~\cite{hu2022lora} of rank 8, together with an adapter implemented accordingly to each baseline operating on hidden-state tokens with a latent dimension of 512.
All models are implemented in PyTorch and trained using GRPO with a group size of 8, a KL coefficient $\beta=0.04$, an effective batch size of 32, and a learning rate of $1.0\times10^{-4}$.
We train for 500 steps on Reasoning Gym benchmarks and 233 steps on GSM8K.
For $\pi_{\text{ref}}$ in \autoref{eq:rlvr_obj}, we insert zero-valued context tokens at the same positions as $\mathcal{C}$ to match the token layout of the policy input.
All experiments are conducted on a single NVIDIA H200 GPU. Each experiment is repeated three times with different random seeds, and we report the mean and standard deviation.
To isolate the effect of expert utilization and control Chain-of-Thought~\cite{kojima2022large} variance, we enforce a fixed generation budget of 128 tokens across all methods including frozen SLM outputs. 
Unless otherwise specified, we fix the number of latent query vectors to $m=8$ and the number of attention heads to 8 for all Perceiver-style cross-attention layers.


\noindent \textbf{LM pool.}
We draw LMs from widely used open instruction-tuned model families that cover complementary skill profiles:
(1) a \textbf{math specialist} Qwen2.5-Math-1.5B-Instruct~\cite{qwen2_5},
(2) a \textbf{coding specialist} Qwen2.5-Coder-7B-Instruct~\cite{qwen2_5},
(3) a \textbf{generalist} Llama-3.2-3B-Instruct~\cite{grattafiori2024llama},
as well as additional general-purpose SLMs including Phi-3.5-mini-instruct~\cite{slm3_abdin2024phi},
gemma-2-2b-it~\cite{team2024gemma},
and Mistral-7B-Instruct-v0.3~\cite{jiang2023mistral7b}.
To stress-test robustness, we also include a \textbf{naive expert}, GPT-2~\cite{radford2019language}.
Finally, we evaluate an alternative large expert GPT-OSS 20B~\cite{gptoss} to test whether the learned interface remains compatible under substantially different expert capacity.
\vspace{-2mm}

\noindent \textbf{Design of Heterogeneous LM Teams.}
\label{subsec:expert_selection}
Our framework is designed to be \textbf{plug-and-play} with respect to the expert team: all experts are treated as frozen, off-the-shelf instruction-tuned LMs, and collaboration is enabled solely through a trainable latent interface.
Thus, we can freely reconfigure the expert team without modifying expert parameters, reflecting a modular perspective where SLMs are orchestrated rather than trained jointly~\cite{slm_future,wu2024autogen,hong2024metagpt}.
In this work, we mainly explore the following expert-team designs:

\noindent
(1) \textbf{Default Team} ($n=3$).
We use a compact team consisting of a math specialist (Qwen2.5-Math-1.5B),
a coding specialist (Qwen2.5-Coder-7B),
and a generalist instruction model (Llama-3.2-3B).
This configuration serves as our primary expert pool in the main experiments.
\vspace{-2mm}

\noindent
(2) \textbf{Generalist Team} ($n=5$).
To test whether our method relies on a specific expert backbone, we additionally consider a larger pool of general-purpose SLMs, including
Phi-3.5-mini,
Gemma-2-2B,
and Mistral-7B,
together with Qwen2.5-Math-1.5B.
This setting emphasizes heterogeneity across model families and capacity.
\vspace{-2mm}

\noindent (3) \textbf{Naive LM Team} ($n=4$) .
We include GPT-2 as a naive (low-capability) expert, alongside Qwen2.5-Math-1.5B, Qwen2.5-Coder-7B, and Llama-3.2-3B.
This stress test evaluates whether the learned interface can suppress low-utility experts rather than exploiting spurious shortcuts.

\vspace{-2mm}

\noindent (4) \textbf{LLM Team} ($n=3$).
Finally, we replace Llama-3.2-3B in the default heterogenous team in (1) with generalist expert with GPT-OSS-20B to test compatibility with substantially different expert capacity.
This verifies that the proposed hidden-state interface remains plug-and-play even under drastic changes in expert scale.

\vspace{-2mm}

\noindent \textbf{Benchmarks.}
We primarily evaluate our method on Reasoning Gym~\cite{reasoninggym}, focusing on task families that prior work has shown to differ in RLVR difficulty.
Specifically, we include Arithmetic and Logic tasks, which are known to be readily solvable by LLMs under RLVR, as well as Algorithmic tasks, which remain challenging and often fail to benefit from RLVR alone.
All tasks are trained using the default settings without curriculum learning, allowing us to isolate the effect of soft hidden state collaboration.
For each Reasoning Gym task family, we train on 16,000 procedurally generated data points and report performance on a held-out validation set of 4,000 samples (Details in \autoref{app:benchmark}).
In addition, we report results on GSM8K~\cite{cobbe2021gsm8k} as a standard mathematical reasoning benchmark to assess whether our method remains competitive in settings where single-model RLVR is already strong.
or GSM8K, we train on the standard training split and evaluate directly on the official test set.

\vspace{-2mm}
\noindent \textbf{Baselines.}
We compare against representative collaboration paradigms while controlling for the downstream RLVR setup.
All methods share the same final policy model, GRPO hyperparameters, and fixed generation budget, and differ only in how expert information is exposed and processed through the context-token construction operator $g_{\phi}$ in \autoref{eq:context_tokens}.
These baselines follow common designs in prior multi-model collaboration and routing-based systems~\cite{debate,routier_r1}.
\vspace{-2mm}

\noindent \textbf{(1) Single-Model RLVR.}
We train the same final policy model ($\pi_{\theta}$) without any auxiliary experts or context tokens, i.e., removing $g_{\phi}$.
This baseline corresponds to standard single-model RLVR training~\cite{reasoninggym}.
\vspace{-2mm}

\noindent \textbf{(2) Output-Level Collaboration} (\autoref{fig:teaser}.a).
Each expert is prompted to generate a short textual hint or solution using the same fixed generation budget of 128 tokens, and the resulting decoded tokens are concatenated verbatim into an \texttt{[Expert Hints]} block.
The operator $g$ maps this text into prefix tokens for the final policy using the tokenizer of the final policy model ($\tau_{\pi}$).
In this setting, $g$ introduces no trainable parameters $\phi$: expert outputs are produced via frozen decoding (including the LM head), and the mapping into context tokens is fixed.
This baseline corresponds to text-mediated collaboration and output aggregation used in debate-style, role-based, and ensembling systems~\cite{debate,llm_blender}:
\vspace{-2mm}
\begin{equation}
g(x)
= \tau_{\pi}\!\left(
[\,\mathrm{decode}_i(LM_i(\tau_i(x)))\,]_{i=1}^n
\right).
\end{equation}
\vspace{-7mm}

\noindent \textbf{(3) Hard Expert Routing (Top-1)} (\autoref{fig:teaser}.b).
We include a hard top-1 expert routing baseline to match router-based expert selection methods~\cite{routier_r1,router_cost_aware_contrastive,router_regret_tsiourvas2025causal},
while omitting task-specific cost modeling and constraint formulations.
Here, $g_{\phi}$ selects a single expert per prompt using a learned router and constructs context tokens exclusively from the selected expert.
Each prompt is first encoded by a sentence-level semantic encoder $\mathcal{E}_{T}$~\cite{wang2020minilm}, whose output is mapped to expert logits by a router head; the expert with the highest logit is selected at evaluation time, while straight-through Gumbel-Softmax is used during training, and we use \texttt{all-MiniLM-L6-v2} as $\mathcal{E}_{T}$.
We perform selective execution by running only the chosen expert and projecting its pooled hidden representation into a single memory token:
\begin{equation}
g_{\phi}(x)
= \mathrm{FFN}\!\left(
\sum_{i=1}^{n} R_i(\mathcal{E}_T(x))\, W_i\, \mathrm{pool}(LM_i(\tau_i(x)))
\right),
\end{equation}
where $R_i(\cdot)\in\{0,1\}$ is the top-1 routing decision over experts, $LM_i$ is the $i$-th frozen expert, and $W_i$ and $\mathrm{FFN}$ are learned projections producing a single memory token.

\noindent \textbf{(4) Ours: Hidden-State Collaboration.}
Frozen experts expose pooled hidden representations, which are aligned and aggregated into $m$ trainable context tokens via a Perceiver-style cross-attention bottleneck~\cite{jaegle2021perceiver}.
Both the latent adapter and context-token construction are optimized end-to-end under RLVR, allowing multiple experts to simultaneously influence the policy through continuous latent representations.

\noindent \textbf{(5) Ours: Without Cross-Attention.}
We remove the Perceiver-style cross-attention and apply the token-wise FFN directly to the stacked expert representations $V$.
This isolates the role of cross-attention in enabling input-dependent expert aggregation.

\begin{equation}
\label{eq:context_tokens_construct_base}
g_{\phi}(x) = \mathrm{FFN}(V).
\end{equation}
\vspace{-4mm}

\vspace{-3mm}
\subsection{Performance analysis}
\vspace{-3mm}
\begin{table}[h]
    \centering
    \caption{Performance comparison across different reasoning tasks on evaluation datasets. We report mean accuracy and standard deviation across 3 runs for all methods and settings considered.}
    \label{tab:performance_comparison}
    \resizebox{\columnwidth}{!}{%
        \begin{tabular}{lcccc}
            \toprule
            Method & Algorithmic & Arithmetic & Logic & GSM8k  \\
            \midrule
            Single & $51.56_{\pm 0.55}$ & $52.34_{\pm 1.78}$ & $\mathbf{96.88}_{\pm1.10}$ & 64.32$_{\pm0.92}$ \\
            Hard Routing & $34.18_{\pm 4.18}$ & $32.47_{\pm 5.10}$ & $63.80_{\pm2.39}$
 & $14.52_{\pm1.52}$\\
            Output Collaboration & $51.43_{\pm3.41}$ & $31.19_{\pm 1.16}$ & $66.80_{\pm0.64}$ & $\mathbf{67.58}_{\pm1.03}$ \\
            Ours (w/o Cross Attn) ($n=3$) & $51.56_{\pm0.64}$
 & $60.16_{\pm4.38}$ & $89.71_{\pm2.17}$ & $63.28_{\pm1.03}$  \\
            \textbf{Ours : Default Team $(n=3)$ } & $51.82_{\pm 0.49}$ & $\mathbf{75.26}_{\pm 5.62}$ & $82.81_{\pm 1.56}$ & $61.59_{\pm1.25}$ \\
            \textbf{Ours : Generalist Team $(n=5)$} & $\mathbf{52.02}_{\pm2.77}$ & $65.10_{\pm8.49}$ & $90.23_{\pm5.01}$ & $41.02_{\pm29.01}$\\
            \bottomrule
        \end{tabular}%
    }
\end{table}
\vspace{-2mm}

\begin{figure*}[ht!]
    \centering
    \includegraphics[width=\linewidth]{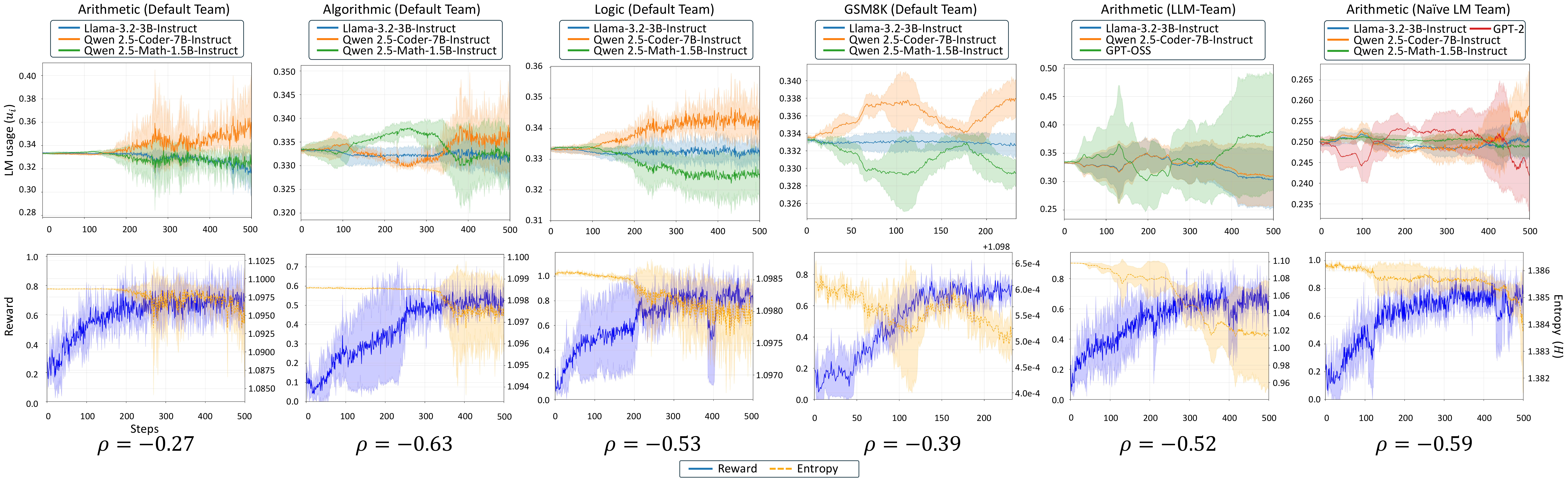}
    \caption{Expert utilization dynamics and training progress under RLVR.
Top row: evolution of per-expert utilization over training steps for different tasks and expert team compositions (Default Team, LLM Team replacement, and Naive Team). Shaded regions indicate standard deviation across runs.
Bottom row: corresponding reward progression and routing entropy over training.
The emergence of selective expert usage without routing supervision suggests that RLVR alone can induce self-organized expert role differentiation.}
    \vspace{-5mm}
    \label{fig:result_expert}
\end{figure*}
\vspace{-2mm}

This section examines whether augmenting a small policy model ($\pi_{\theta}$) with additional trainable context tokens ($\mathcal{C}$) yields consistent performance gains under RLVR.
As shown in \autoref{tab:performance_comparison}, the effect of latent expert conditioning is highly regime-dependent.
On Arithmetic, we observe a clear and consistent improvement, where the full model substantially outperforms the Single-model baseline.
This indicates that expert-derived latent context can provide a useful inductive bias in arithmetic-dominant settings, particularly when the RLVR objective leaves substantial capacity for further performance improvement and the base policy alone is insufficient.
We also observe substantially increased variance across runs in our proposed framework, especially for larger expert teams, indicating that while our method can achieve strong performance, it is sensitive to training dynamics and alignment with the latent expert representations.

In contrast, improvements on Algorithmic tasks are negligible.
Even with latent expert context, performance remains close to the Single-model baseline, indicating that expert conditioning does not fundamentally resolve the known difficulty of algorithmic reasoning under RLVR.
This behavior is consistent with prior observations reported in \cite{reasoninggym}.
For Logic and especially GSM8K, adding expert context provides limited benefit and often degrades performance.
These tasks are already well-handled by a single SLM under RLVR, and the Single-model operates near saturation.
In such regimes, additional context tokens introduce auxiliary latent variables that offer little new information and instead act as structured noise, interfering with an otherwise sufficient internal reasoning process.

Taken together, these results suggest that latent context tokens are not universally beneficial under RLVR.
Their effectiveness is tightly coupled to task difficulty and the remaining capacity for improvement under the training objective.
When the task remains challenging, latent expert representations can inject complementary structure that improves performance.
However, for relatively simple or saturated tasks, such as GSM8K, multi-expert conditioning becomes unnecessary and can be counterproductive.
This highlights a practical constraint: hidden state collaboration should be applied selectively, rather than treated as a general-purpose improvement over single-model RLVR.
Across tasks, we observe that failures to exploit SLM information often manifest as high run-to-run variance, reflecting the difficulty of reliably coordinating multiple experts under RLVR.

\subsection{Does RLVR discover LM expert roles?}

This section investigates whether RLVR induces structured and task-dependent expert utilization patterns among frozen language models, despite the absence of any supervision on routing decisions or expert identities.
In our framework, the only observable signal of expert selection is the latent cross-attention weight matrix $A \in \mathbb{R}^{m \times n}$, where each row represents a distribution over the $n$ experts for one latent query.
To characterize expert selectivity and its relation to learning dynamics, we track routing entropy $H$, per-expert utilization $u_i$, and their association with reward using Spearman rank correlation computed over training trajectories, i.e., between the time series of reward and routing entropy during RLVR:
\begin{equation}
\begin{aligned}
H \;&=\; \frac{1}{m}\sum_{j=1}^{m}\Big(-\sum_{i=1}^{n} A_{j,i}\log A_{j,i}\Big), \quad 
u_i = \frac{1}{m}\sum_{j=1}^{m} A_{j,i}.
\end{aligned}
\end{equation}
Lower entropy indicates more concentrated expert usage.
All statistics are averaged over the batch, and correlations are computed per run and then averaged.

As shown in \autoref{fig:result_expert}, RLVR alone is sufficient to induce non-trivial expert differentiation.
Across tasks, routing evolves from an initially uniform mixture into structured allocations that concentrate attention on a subset of consistently useful models.
This process is reflected by a monotonic decrease in routing entropy and a consistent negative correlation between entropy and reward (mean Spearman $\rho = -0.48$), indicating that improvements in reward over training co-occur with increasingly selective expert usage.
Importantly, the resulting structure adapts to the expert pool: weak experts such as GPT-2 are suppressed throughout training, whereas introducing a stronger expert such as GPT-OSS leads to a pronounced redistribution of attention toward it.

These dynamics admit two complementary interpretations.
One is functional, where the adapter discovers reward-relevant expert roles based on representational utility.
The other is an implicit capacity bias, where higher-capacity models are favored irrespective of nominal domain specialization, as suggested by the dominance of GPT-OSS and the preference for Qwen-Coder over Qwen-Math even on arithmetic tasks.
Although expert representations are projected into a shared space before cross-attention, the expert-specific projections $W_i$ in \autoref{eq:dim_adapter} scale with the original hidden dimensionality, which may introduce a weak inductive advantage.
Thus, while RLVR clearly induces robust, self-organized expert utilization without metadata or routing supervision, disentangling functional expertise from capacity-driven preference remains an open question.

\begin{figure*}
\vspace{-3mm}
\centering
\begin{minipage}{0.63\linewidth}
    \centering
    \includegraphics[width=\linewidth]{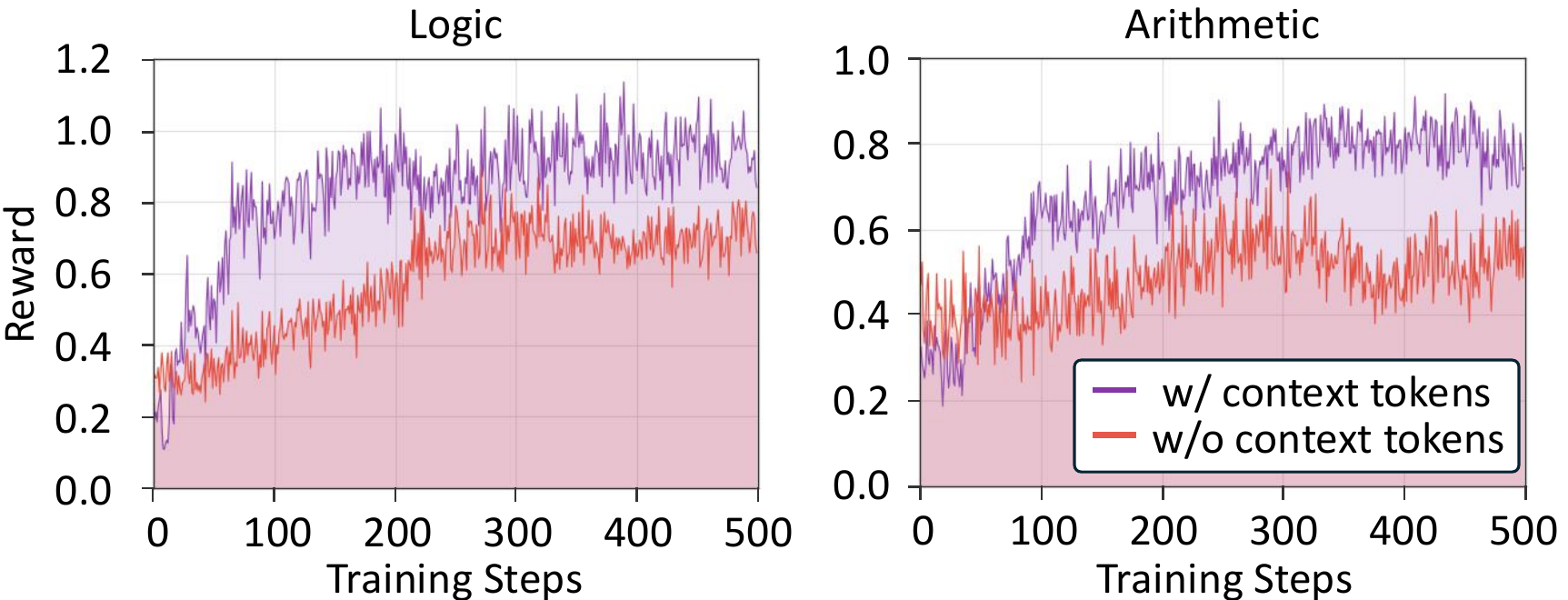}
    \vspace{-6mm}
    \caption{Ablation comparing the proposed model (with $\mathcal{C}$) and a baseline without $\mathcal{C}$ on Logic and Arithmetic. Models with $\mathcal{C}$ achieve consistently higher rewards.}
    \label{fig:wo_token}
    \vspace{-6mm}
\end{minipage}
\hfill
\begin{minipage}{0.34\linewidth}
    \centering
    \includegraphics[width=\linewidth]{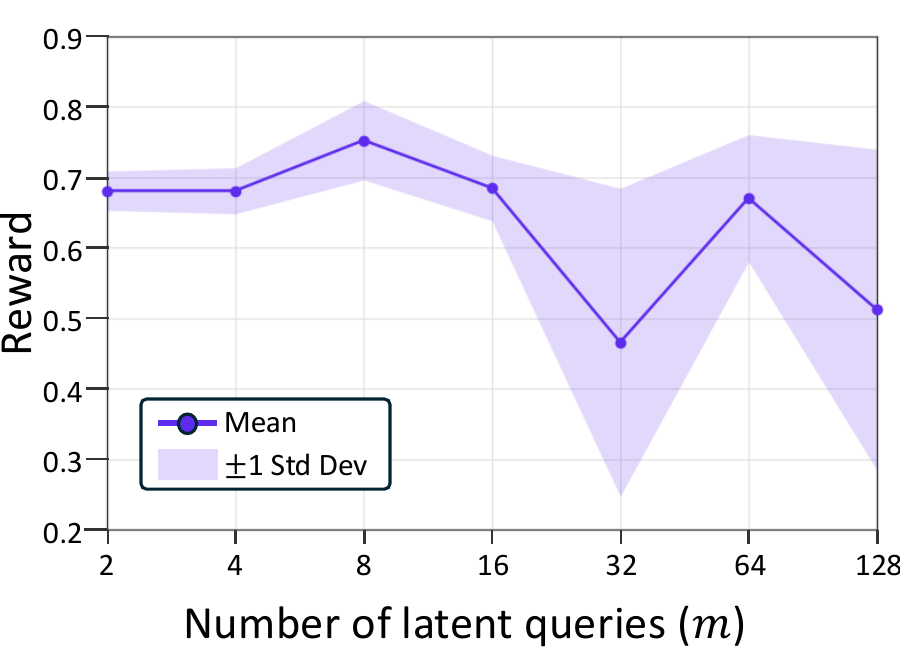}
    \vspace{-6mm}
    \caption{Effect of the number of latent queries $m$ on reward performance.}
    \label{fig:latent}
    \vspace{-6mm}
\end{minipage}
\end{figure*}

\vspace{-3mm}
\subsection{Ablation Studies}
\vspace{-2mm}
\begin{table}[h]
    \centering
    \caption{Ablation study on expert hidden-state pooling strategies (Arithmetic). We report mean accuracy and standard deviation.}
    \label{tab:ablation_pooling}
    \resizebox{\columnwidth}{!}{%
        \begin{tabular}{lccccc}
            \toprule
            & First Token ($h_{i, 0}^{(L_{i})}$) & First Layer($h_{i, T_{i}}^{(0)}$) & Random Token & Random Layer & Ours ($n=3$) \\
            \midrule
            Accuracy (\%)
            & $75.13_{\pm 1.12}$
            & $54.62_{\pm 17.82}$
            & $63.93_{\pm 6.33}$
            & $38.09_{\pm25.55}$
            & $\mathbf{75.26}_{\pm 5.62}$ \\
            \bottomrule
        \end{tabular}%
    }
    \vspace{-3mm}
\end{table}
\vspace{-1mm}
\noindent \textbf{Do context tokens $\mathcal{C}$ contribute to SLM reasoning?}
To evaluate whether the $\mathcal{C}$ provide additional conditioning beyond the input $x$, we compare the same policy $\pi_{\theta}$ under two generation settings during training:
conditioning on the input alone, $\pi_{\theta}(y \mid x)$, and conditioning on the input augmented with context tokens, $\pi_{\theta}(y \mid [x \,\|\, \mathcal{C}])$.
As shown in~\autoref{fig:wo_token}, conditioning on $\mathcal{C}$ yields consistently higher rewards throughout training.
In Logic, the model conditioned on $\mathcal{C}$ reaches higher reward within 500 training steps, whereas conditioning on $x$ alone saturates at a substantially lower level.
A similar trend is observed in Arithmetic, where context-conditioned generation leads to higher rewards across the entire training trajectory. These results indicate that the learned context tokens $\mathcal{C}$ make a non-trivial contribution during training, rather than performance improvements arising solely from optimizing $\pi_{\theta}$.




\noindent \textbf{Disentangling Token Position and Depth.}
We examine whether the default choice of last-token pooling from the final layer in \autoref{eq:last_token} is essential by conducting pooling ablations on the Arithmetic subset that vary token position and depth. Specifically, we replace last-token pooling with first-token pooling, uniformly sampled token pooling, first-layer pooling, and uniformly sampled layer pooling. As shown in \autoref{tab:ablation_pooling}, first-token pooling achieves performance comparable to the default configuration, while pooling from shallow layers or uniformly sampled layers leads to unstable or degraded performance. Uniformly sampling a token position remains competitive but exhibits higher variance, suggesting that precise token alignment is not critical. Overall, these results may indicate that the proposed method does not rely on fine-grained, prompt-conditioned reasoning traces within SLM outputs; instead, a coarse and relatively prompt-invariant representation of each SLM is sufficient, with expert hidden states acting as stable model signatures that regularize the policy under RLVR.

\vspace{-2mm}
\noindent \textbf{Latent dimension analysis.}
We study the effect of the number of latent queries $m \in \{2,4,8,16,32,64,128\}$ on performance.
As shown in \autoref{fig:latent}, performance is maximized at moderate latent sizes ($m=8$) and degrades as $m$ increases further.
Larger latent dimensions also exhibit higher variance and less stable training, indicating that latent bottleneck capacity does not translate monotonically into better performance.

\vspace{-2mm}
\section{Related Works}
\vspace{-2mm}

\vspace{-1mm}
\noindent \textbf{LLMs as Collaborative Agents:}
Following prior work that highlights the role of SLMs as building blocks for agentic systems~\cite{slm_future}, we use the term SLM to refer to relatively compact LMs below roughly 10B parameters.
Recent studies have demonstrated that strong performance can be achieved with such models through a variety of approaches, including efficient hybrid architectures~\cite{slm1_fu2025nemotronflash,slm_5_blakeman2025nemotron}, retrieval-augmented modeling~\cite{slm2_retro_pmlr-v162-borgeaud22a}, and data-centric or RL-based training strategies~\cite{slm3_abdin2024phi,slm4_allal2025smollm2,guo2025deepseek}.
Together, these advances support a growing view that collections of specialized LMs can serve as effective, modular building blocks for agentic systems~\cite{su2025difficulty,yuan2025automated,zhang2026agentorchestra}.
This trend toward modularity is further exemplified by emerging standards like the Model Context Protocol (MCP)~\cite{mcp2024anthropic} and large-scale API-tuning frameworks~\cite{patil2024gorilla, qin2024toolllm}, which aim to unify how models interface with external tools via discrete text or JSON.
Whereas prior work has primarily examined such collaboration at the system or orchestration level, our interest lies at a different granularity: whether and how collaboration can emerge at the model level, through shared learning signals and internal representations, and what this perspective reveals about individual experts.

\vspace{-2mm}
\noindent \textbf{Coordination Paradigms for Heterogeneous LMs.} Prior work on multi-LM collaborations typically follows three paradigms: expert routing for efficient selection under constraints~\cite{routier_r1,router_cost_aware_contrastive,router_regret_tsiourvas2025causal, selection1, selection2, selection3}, output-level collaboration via textual interaction or ensembling~\cite{debate,wu2024autogen,qian2024chatdev}, and architectural-level MoE~\cite{switchtransformer,heteromoe} that yields structured specialization~\cite{router_cook2025brainlike}. While routing optimizes invocation, it provides limited mechanisms for joint expert utilization once models are available. Similarly, decoding-time interfaces—including distribution fusion in DeePEn~\cite{huang2024ensemble} and collaborative decoding in Co-LLM~\cite{co-llm}—can be viewed as non-architectural variants of MoE that operate on frozen experts. 
However, by interfacing at the output level, these methods primarily rely on emitted distributions or token-level routing decisions~\cite{huang2024ensemble,co-llm}. This coarse granularity overlooks evidence that intermediate hidden states can provide richer task-relevant features than final-layer outputs~\cite{skean2025layer}. 
Consequently, we shift collaboration to the representation level by interfacing through latent hidden states.
This moves coordination away from emitted tokens or distributions and instead exposes intermediate expert representations directly to the policy.
While the interface remains latent, this finer-grained design provides a principled way to observe how expert utilization evolves under RLVR, allowing multiple frozen LMs to influence the policy simultaneously without the pretraining costs of architecture-level MoE, and revealing emergent roles for each LM.

\vspace{-5mm}
\section{Conclusions}
\vspace{-3mm}

We show that RLVR induces emergent specialization among frozen LM experts through a simple adapter-based soft prompt tuning mechanism, without any explicit supervision on routing decisions or expert roles.
Compared to hard routing and output-level fusion, this formulation avoids brittle decisions and token inefficiency while enabling multiple experts to influence the policy simultaneously at the representation level.
Under RLVR, expert utilization evolves from uniform attention to structured, non-uniform allocation aligned with reward improvement, indicating that role differentiation can emerge purely from outcome-level supervision.
At the same time, the benefits of soft prompt--based collaboration are task-dependent: on simpler benchmarks, single-model RLVR already saturates performance and additional experts can introduce interference, revealing a trade-off between single-model sufficiency and multi-expert conditioning.
We note that our hard-routing comparison does not exhaust the space of possible router designs, and that our framework prioritizes understanding expert utilization under outcome-level supervision rather than computational efficiency.
Future work includes exploring a wider range of agent compositions and interaction patterns, incorporating selective execution or capacity-matched routing baselines, and revisiting this comparison as SLMs continue to shrink, where finer-grained specialization and more pronounced trade-offs may become easier to study.


\normalsize
\bibliography{references}

\newpage

\onecolumn

\section*{Appendix}

\section{Additional Implementation Details}
\label{app:perceiver_adapter}

\subsection{Feed-Forward Network}
\label{app:ffn}

The token-wise feed-forward network (FFN) following the cross-attention operation in \autoref{eq:context_tokens_construct} is implemented as a two-layer multilayer perceptron with a bottleneck structure:
\begin{equation}
\mathrm{FFN}(x) = \tanh (W_2 \, \tanh\!\left(W_1 x \right)),
\end{equation}
where $W_1 \in \mathbb{R}^{d \times d/2}$ and $W_2 \in \mathbb{R}^{d/2 \times d}$. 
We use $\tanh$ as the nonlinearity throughout all experiments.

\section{Reasoning Gym Benchmark Details.}
\label{app:benchmark}

This section summarizes the benchmark configurations used in our experiments.
All Reasoning Gym settings follow the default settings of the Reasoning Gym~\cite{reasoninggym} framework.
We include representative YAML snippets for general training hyperparameters and for each task family to facilitate reproducibility.

\begin{lstlisting}[language=yaml, caption={General Trainig Settings}, label={lst:train_config}]
data:
  train_batch_size: 4
  val_batch_size: 4
  
trainer:
  is_sft: false
  multiple_lr: true
  lr: 1e-4   # LR for policy (actor)
  router_lr: 1e-4
  epsilon: 1e-6
  weight_decay: 1e-2
  total_epochs:  1
  total_training_steps: 500 # 4 * 8 * 500 = 16,000 data points
  log_freq: 8
  save_freq: 1000
  test_freq: 1000
  eval_strategy: "steps" 
  eval_batches: 1000 #(batch size) * 1,000 = 4,000 validation data points.
  grad_clip: 1.0
  warmup_steps: null
  gradient_accumulation_steps: 8 # Thus the net batch size is 32.
\end{lstlisting}

\begin{lstlisting}[language=yaml, caption={Training configuration for Arithmetic}, label={lst:arithmetic}]
reasoning_gym:
  dataset_size: 20000

  developer_prompt: DeepSeekZero
  datasets:
    complex_arithmetic:
        weight: 1
    intermediate_integration:
      weight: 1
    polynomial_equations:
      weight: 1
    polynomial_multiplication:
      weight: 1
    simple_geometry:
      weight: 1
    bitwise_arithmetic:
      weight: 1
    chain_sum:
      weight: 1
    decimal_arithmetic:
      weight: 1
    decimal_chain_sum:
      weight: 1

curriculum:
    enabled: False
    schedule:
      automatic: True
      update_steps: 30 # automatic curriculum updating after 50 steps
    last_k: 20
    success_threshold: 0.70
    failure_threshold: 0.10
    curricula:
      spell_backward:
        attribute_levels:
          word_len: 0

\end{lstlisting}

\begin{lstlisting}[language=yaml, caption={Training configuration for Algorithmic}, label={lst:algo}]
reasoning_gym:
  dataset_size: 20000
  developer_prompt: DeepSeekZero
  datasets:
    ab:
      weight: 1
    base_conversion:
      weight: 1
    binary_alternation:
      weight: 1
      config:
        p_solvable: 0.9
    binary_matrix:
      weight: 1
      config:
        min_n: 2
        max_n: 6
    caesar_cipher:
      weight: 1
      config:
        max_words: 10
    cryptarithm:
      weight: 1
    isomorphic_strings:
      weight: 1
      config:
        max_string_length: 8
    jugs:
      weight: 1
      config:
        difficulty: 6
    rotate_matrix:
      weight: 1
      config:
        min_n: 2
        max_n: 6
    string_manipulation:
      weight: 1
      config:
        max_string_length: 15
        max_num_rules: 6

curriculum:
    enabled: False
    schedule:
      automatic: True
      update_steps: 30 # automatic curriculum updating after 50 steps
    last_k: 20
    success_threshold: 0.70
    failure_threshold: 0.10
    curricula:
      spell_backward:
        attribute_levels:
          word_len: 0

\end{lstlisting}

\begin{lstlisting}[language=yaml, caption={Training configuration for Logic}, label={lst:logic}]
reasoning_gym:
  dataset_size: 20000
  developer_prompt: DeepSeekZero
  datasets:
    aiw:
      weight: 1
    circuit_logic:
      weight: 1
    knights_knaves:
      weight: 1
    propositional_logic:
      weight: 1
    self_reference:
      weight: 1
    syllogism:
      weight: 1
    zebra_puzzles:
      weight: 1

curriculum:
    enabled: False
    schedule:
      automatic: True
      update_steps: 30 # automatic curriculum updating after 50 steps
    last_k: 20
    success_threshold: 0.70
    failure_threshold: 0.10
    curricula:
      spell_backward:
        attribute_levels:
          word_len: 0


\end{lstlisting}


\end{document}